\documentclass[10pt, a4paper, table]{article}
\usepackage[final]{lrec-coling2024} 
\usepackage{booktabs}
\usepackage[nameinlink]{cleveref}
\usepackage{xurl}
\usepackage{multirow}
\usepackage{booktabs, multirow} 
\usepackage{soul}
\usepackage{changepage,threeparttable} 
\title{Sign of the Times: Evaluating the use of Large Language Models for Idiomaticity Detection}

\name{Dylan Phelps\textsuperscript{1,2}, Thomas Pickard\textsuperscript{2}, Maggie Mi\textsuperscript{2}, Edward Gow-Smith\textsuperscript{2}\\ 
{\bf \large Aline Villavicencio\textsuperscript{2,3}}
}

\address{
    \textsuperscript{1}Healthy Lifespan Institute, The University of Sheffield, United Kingdom\\
    \textsuperscript{2}Department of Computer Science, The University of Sheffield, United Kingdom\\
    \textsuperscript{3}Institute for Data Science and Artificial Intelligence, The University of Exeter, United Kingdom\\
    \{drsphelps1, tmpickard1, zmi1, egow-smith1\}@sheffield.ac.uk\\
    a.villavicencio@exeter.ac.uk\\
}

\abstract{
Despite the recent ubiquity of large language models and their high zero-shot prompted performance across a wide range of tasks, it is still not known how well they perform on tasks which require processing of potentially idiomatic language. In particular, how well do such models perform in comparison to encoder-only models fine-tuned specifically for idiomaticity tasks? In this work, we attempt to answer this question by looking at the performance of a range of LLMs (both local and software-as-a-service models) on three idiomaticity datasets: SemEval 2022 Task 2a, FLUTE, and MAGPIE. Overall, we find that whilst these models do give competitive performance, they do not match the results of fine-tuned task-specific models, even at the largest scales (e.g. for GPT-4). Nevertheless, we do see consistent performance improvements across model scale. Additionally, we investigate prompting approaches to improve performance, and discuss the practicalities of using LLMs for these tasks. 
 \\ \newline  }

\begin{document}

\maketitleabstract

\section{Introduction}
Large, pre-trained language models (LLMs) are becoming increasingly popular in academic, industrial, and lay spheres due to their ability to perform well across a range of tasks in a zero-shot or few-shot prompting set-up, including question answering, common-sense reasoning \cite{openai2023gpt4, geminiteam2023gemini}, and machine translation \cite{xu_paradigm_2023,koshkin_transllama_2024,dabre_study_2023}. Despite this, there is yet to be an analysis of how well such models are able to handle potentially idiomatic language.
Much previous work has shown that smaller, encoder-only transformer models have poor performance in identifying and representing idiomatic expressions when pre-trained on a large general dataset \cite{nandakumar2019well, garcia2021probing}. However, the performance of such models increase hugely when they are fine-tuned on a task-specific dataset containing a large number of idiomatic expressions \cite{madabushi2021astitchinlanguagemodels, zeng_idiomatic_disc}. This fine-tuning procedure, however, requires dedicated hardware and training, something that isn't possible with LLMs on an academic budget.

In this work, we benchmark the performance of several widely-used LLMs (using both software-as-a-service remote implementations and local instances) on three in-context idiomaticity detection datasets; the idiom portion of FLUTE \citelanguageresource{chakrabarty2022flute}, MAGPIE \citelanguageresource{haagsma2020magpie}, and SemEval 2022 Task 2a \cite{tayyar-madabushi-etal-2022-semeval}. FLUTE and MAGPIE cover English (EN) only, while the SemEval dataset also includes expressions in Brazilian Portuguese (PT-BR) and Galician (GL).

Overall, our experiments show that large LLMs give competitive performance on idiomaticity datasets, which can be generally applied due to the lack of type specific fine-tuning, but nevertheless lag in general behind much-smaller finetuned encoder-only models. We also find that idiomaticity detection performance still scales with the number of parameters in the model. Finally, we discuss a number of considerations affecting the models' performance and the practicality of using them for idiomaticity detection, including the training dataset and the capability of the model to follow instructions given in the prompt.

\vspace{-1em}
\section{Datasets}
We investigate the performance of LLMs on three datasets consisting of potentially idiomatic expressions in context. The datasets are chosen to provide a diverse set of potentially idiomatic expressions which feature a range of morphological forms and variations across two different tasks: textual entailment and idiomaticity detection. 1,859 different English target expressions are represented across the three datasets.
We focus on English, but the inclusion of Semeval 2022 Task 2a allows us to additionally explore performance across languages.

\subsection{FLUTE}

FLUTE \citelanguageresource{chakrabarty2022flute} frames the understanding of four kinds of figurative language (sarcasm, simile, metaphor and idioms) as a natural language inference (NLI) task, in which pairs of literal and figurative sentences are labelled as either entailing or contradicting one another.
The sentence pairs are generated using a model-in-the-loop approach, with base text generated by GPT-3 which is then edited by crowdworkers and reviewed by experts.

For our analysis, we consider only the idiom section of the FLUTE dataset, which consists of 1,768 training examples across 479 idioms and a further 250 test examples across 69 idioms. No idiom appears in both the training and test sets.

\citealplanguageresource{chakrabarty2022flute} provide benchmark performance metrics using T5 models \cite{raffel2020exploring} on the FLUTE training data, reporting 79.2\% accuracy (0.791 macro-F1).
A FigLang22 shared task using the FLUTE dataset \cite{saakyan_report_2022} attracted several entries, with the best-performing systems developed by \cite{gu_just-dream-about-it_2022} and \cite{bigoulaeva_effective_2022}. The latter adopt a pipeline approach, improving the T5 baseline by sequentially fine-tuning on e-SNLI dataset \citelanguageresource{e-SNLI} and IMPLI (which incorporates figurative language) \citelanguageresource{stowe_impli_2022}, followed by the task dataset. Using the authors' published outputs, we calculate a macro-average F1 of 0.952 on the idiom portion of the FLUTE test set.

\subsection{SemEval 2022 Task 2a}

SemEval 2022 Task 2a \cite{tayyar-madabushi-etal-2022-semeval} is a binary classification idiomaticity detection task, in which a potentially idiomatic noun compound, as used in a given context sentence, must be labelled as either literal or idiomatic. The dataset includes compounds across a range of idiomaticity, including fully compositional (\textit{insurance company}) as well as partially (\textit{eager beaver}) and entirely opaque (\textit{sugar daddy}) items.
The task offers both ``one-shot" and ``zero-shot" settings; the former is evaluated with new context instances of previously-seen items, while the latter uses compounds not present in the training data for evaluation. 

The test set for the task contains 50 compounds each in English (with 916 instances), Brazilian Portuguese (713 instances) and Galician (713 instances).

\Cref{table:semeval_refscores} shows the macro-F1 scores in the zero-shot and one-shot settings for the baseline models \cite[fine-tuned multilingual mBERT, per][]{madabushi2021astitchinlanguagemodels} and the best-performing entries to the shared task\footnote{For the one-shot setting, the best-performing model is a fine-tuned multilingual XLM-RoBERTa, as described in \citet{semeval_hit}.}. 

\begin{table}[h!]
\centering 
\small
\resizebox{\columnwidth}{!}{
\begin{tabular}{llcccc}
\toprule
 \multirow{3}{*}{Setting} & \multirow{3}{*}{Reference} & \multicolumn{4}{c}{Language} \\
 \cline{3-6}
\vspace{-0.9em}
  & & & & & \\
  & & EN & PT & GL & All \\
\midrule
 \multirow{2}{*}{Zero-Shot} & Best & 0.902 & 0.828 & 0.928 & 0.890 \\
  & Baseline & 0.707 & 0.680 & 0.507 & 0.654 \\
  \cline{1-6}\vspace{-0.8em}\\
  \multirow{2}{*}{One-Shot} & Best & 0.964 & 0.894 & 0.937 & 0.939 \\
  & Baseline & 0.886 & 0.864 & 0.816 & 0.865 \\
\bottomrule
\end{tabular}
}
\caption{\label{table:semeval_refscores} \centering Reference scores (Macro F1) for SemEval 2022 Task 2a.}
\end{table}

\subsection{MAGPIE}

MAGPIE \citelanguageresource{haagsma2020magpie} is a corpus of instances of potentially idiomatic expressions (PIEs -- expressions which have multiple senses, including at least one with a high level of idiomaticity), in which each instance has been annotated as either idiomatic, literal, or other (proper noun, etc.) by a group of crowd-sourced workers. The PIEs in the dataset are chosen from three online dictionaries and so have a wide range of forms and frequencies.

The final dataset consists of 56,622 annotated instances, of which ~70\% are idiomatic, ~28\% are literal and ~1\% are other. In our experiments we use the test split of the randomly split dataset, which has 4,840 instances across 1,134 PIEs).

\citetlanguageresource{haagsma2020magpie} do not provide baseline models for the MAGPIE data, but several benchmarks are provided by \citet{zeng_idiomatic_disc}.

\subsection{Construction Artifacts}

Recent work by \citet{boisson_constructionartifacts} has found that language models tuned for metaphor identification (in which they include idiomaticity detection) on artificially-constructed datasets (i.e. those not sampled from `naturally-occurring' text) can perform well when the target expression or the surrounding context are hidden from the model, ``in both cases close to the model with complete information".

As our experiments employ pre-trained LLMs without fine-tuning for the idiomaticity detection task, we anticipate that the concerns highlighted by \citet{boisson_constructionartifacts} should not affect our findings. 
While the training regimes for many of the models we examine are not public, it seems likely that they have consumed large quantities of training data containing `naturally distributed' idiomatic expressions. 

It is also worth noting that we can not rule out the possibility that these LLMs' training data includes the training or test datasets under evaluation\footnote{The SemEval test set is publicly available only without labels; FLUTE and MAGPIE are public.}, and it is likely (for SemEval and MAGPIE) that the context sentences could have been `seen' by the models during training (albeit without idiomaticity markers), as they are taken from online sources.

\section{Models}

To be able to compare results from a range of currently-available LLMs, we evaluate both software-as-a-service (SaaS) and local instances of open models. To maximise applicability of our findings to researchers, we focus on local instances that can be run on consumer-level hardware (targeting a machine with 32GB RAM and 12GB VRAM).

\Cref{table:model_comparison} summarises the models used in our experiments, including the parameter count (where available), cost to run for SaaS models, and whether the training dataset is multilingual.

\begin{table}[h!]
\centering  
\resizebox{\columnwidth}{!}{
\small
\begin{tabular}{rccc}
\toprule
Model & Params (billions) & Cost (\$US per 1000 tokens) & Multilingual  \\
\midrule
GPT-3.5-turbo & Unknown & 0.0005 & Y \\ 
GPT-4-turbo & Unknown & 0.01 & Y  \\ 
GPT-4 & Unknown & 0.03 & Y \\ 
Gemini-1.0 Pro & Unknown & 0.000125 & Y \\ 
Llama2-7B-chat & 7 & N/A & N \\ 
Llama2-13B-chat & 13 & N/A & N \\ 
Llama2-70B-chat & 70 & N/A & N \\ 
Phi-2 & 2.5 & N/A & N \\ 
Mistral-7B & 7 & N/A & N \\
Flan-T5-Small &0.08 &N/A &Y\\
Flan-T5-Base &0.25 &N/A &Y\\
Flan-T5-Large &0.78 &N/A &Y\\
Flan-T5-XL &3 &N/A &Y\\
Flan-T5-XXL &11 &N/A &Y\\
\bottomrule
\end{tabular}
}
\caption{\label{table:model_comparison} \centering  Characteristics of the models evaluated.}
\end{table}

\subsection{Software-as-a-service Models}

\subsubsection{OpenAI}

OpenAI models are seen to be the current state of the art in SaaS models. GPT-4 \cite{openai2023gpt4}, their current largest model, has been shown to achieve or exceed human-level performance in a number of commonly used benchmarks. We evaluate GPT-3.5-turbo (gpt-3.5-turbo-0613), GPT-4-turbo (gpt-4-0125-preview) and GPT-4 (gpt-4) in this work. GPT-3.5 is a smaller model created as a test run during the development of GPT-4, and GPT-4-turbo is an optimised and more recent variant of GPT-4. The parameter counts for these models are not known, but it is assumed that GPT-4 is substantially larger than GPT-3.5.


\subsubsection{Google}

Google provides access to a number of models of varying size and price through its VertexAI API. In this work we evaluate the performance of the Gemini Pro 1.0 model. Gemini Pro is trained on a multimodal and multilingual dataset and its performance exceeds that of GPT-3.5 on a number of benchmarks \cite{geminiteam2023gemini}.

\subsection{Local Models}

Additionally, we evaluate the performance of popular open models that can be run locally. The models chosen are the Llama2 models, \cite{touvron2023llama} Llama2-7B-chat and Llama2-13B-chat, Phi-2 \cite{li2023textbooks, phi2}, and the CapybaraHermes\footnote{\url{https://huggingface.co/argilla/CapybaraHermes-2.5-Mistral-7B}} variant of Mistral-7B \cite{jiang2023mistral}.

To ensure that the models can be run on consumer-level hardware we use quantized variants of each model with 7B or more parameters. Quantization \cite{dettmers2022llmint8, frantar2023gptq} involves converting each parameter from full 16-bit floating point numbers to a set of $2^n$ discrete values. This massively reduces the size of the models so they can be run on a wider range of hardware, with a trade-off of lower performance. We use \textbf{Q5\_K\_S quantisation variants}, which use 5-bit quantization, provided by TheBloke on Huggingface\footnote{\url{https://huggingface.co/TheBloke}}. 5 bit quantization has been shown to have minimal impact on the performance of the model\footnote{See \url{https://github.com/ggerganov/llama.cpp/pull/1684}.}.

To run the models we use the Huggingface transformers library \cite{wolf2020huggingfaces} for Phi-2 and llama.cpp\footnote{\url{https://github.com/ggerganov/llama.cpp}} for all the quantized models.

\subsection{Multilingual Models}
We also explore the performance of multilingual models. In particular, we target our exploration to variants of the Flan-T5 models \citep{chung2022scaling}: Flan-T5-Small, Flan-T5-Base, Flan-T5-Large, Flan-T5-XL, and Flan-T5-XXL.

We are interested in how multilingual models' performance on idiomatic language-related tasks differs from monolingual ones. Moreover, we want to investigate the extent to which the performance is impacted by model size.

\begin{table*}[h]
\centering  
{
\begin{tabular}{lccc}
\toprule
 & SemEval & FLUTE & MAGPIE \\
\midrule
GPT-3.5-Turbo & 0.645 & 0.820 & 0.559 \\
GPT-4-turbo & 0.668 & 0.936 & 0.860 \\
GPT-4 & 0.636 & 0.936 & 0.896 \\
Gemini 1.0 Pro & 0.672 & 0.924 & 0.721 \\
\midrule
Phi-2 & 0.447 & 0.458 & 0.531 \\
Llama2 (7B-chat) & 0.479 & 0.373 & 0.314 \\
Llama2 (13B-chat) & 0.505 & 0.602 & 0.483 \\
CapybaraHermes-2.5-Mistral-7B & 0.539  & 0.812 & 0.587 \\
\midrule
Flan-T5-Small &0.333 &0.333 &0.203 \\
Flan-T5-Base &0.390 &0.764 &0.213 \\
Flan-T5-Large &0.424 &0.872 &0.290 \\
Flan-T5-XL &0.452 &0.956 &0.456 \\
Flan-T5-XXL (11.3B) &0.514 &0.940 &0.753 \\
\midrule
\textit{baseline} & 0.654 & 0.791 & 0.872 \\
\textit{best} & 0.890 & 0.952 & 0.955 \\
\bottomrule
\end{tabular}
}
\caption{\label{table:results_full} \centering Main results of our models across the three idiomaticity datasets. All results presented are macro-average F1 scores over the two classes. Baseline results are taken from \citet{madabushi2021astitchinlanguagemodels}, \citet{chakrabarty2022flute} and \citet{zeng_idiomatic_disc}. `Best' results (in all cases using models fine-tuned on the task training data) are taken from \citet{semeval_hit}, \citet{bigoulaeva_effective_2022} and \citet{zeng_idiomatic_disc}. For SemEval, the `zero-shot' setting is reported.}
\end{table*}

\section{Results}

Our main results across the three datasets (using our default prompts) are shown in \Cref{table:results_full}. 
To make our results representative and generalisable, we ran the models multiple times, where not computation or cost prohibitive -- all of the Flan models were run three times, whilst the Gemini Pro and GPT-3.5 models were run twice on SemEval, which is particularly important for reducing the variance of the results when testing different prompting methods; all other models were run once only.

Comparing the results with the baseline and best-performing models, we can see that while the performance of large, contemporary LLMs may be higher than out-of-the-box encoder-only models, there is still a gap between them and the results which can achieved by encoders fine-tuned to the particular tasks. However, given the work of \citet{boisson_constructionartifacts} on construction artifacts within datasets for idiomaticity detection, the ability of LLMs to disambiguate a wide-range of PIEs without additional fine-tuning shows the general ability of these models to detect idiomaticity, which may not have been achieved by fine-tuned encoders.

\subsection{Model Scaling}
\begin{figure}[h!]
\centering
\includegraphics[width=0.9\columnwidth]{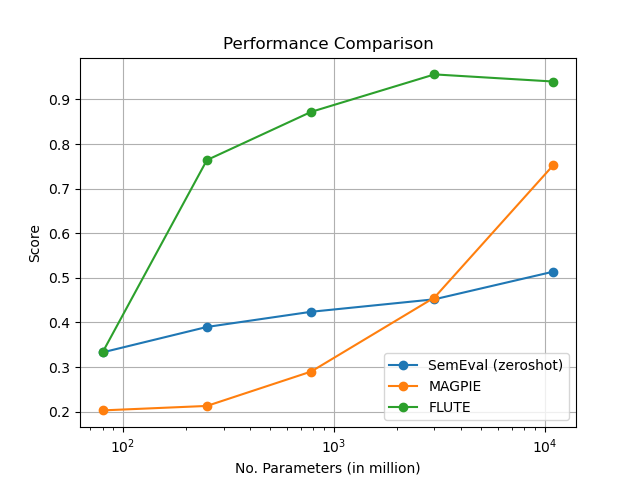}
\caption{\centering Performance on the three datasets for different Flan-T5 model sizes.}
\label{fig:performance_vs_params}
\end{figure}

With the exception of the Mistral-7B model, there is a significant gap in performance between the smaller, locally-run models and the larger SaaS models. We can also see the same trend for our Llama2 models, where the larger Llama2-13B model outperforms the smaller Llama2-7B one on all datasets and splits.
From the results of the Flan-T5 model variants, as shown in \Cref{fig:performance_vs_params}, there is a clear trend that increasing model size leads to improved performance. 
This trend appears to slow down somewhat after model size reaches around 3B parameters (Flan-T5-XL), though performance on the MAGPIE dataset continues to grow.


\subsection{Prompts}
Due to the differing input formats required by the various models, we use slightly different prompts. 
Here, we show our default prompts used for the GPT models. For SemEval and MAGPIE, we use: 

``Disambiguate whether the given expression is used idiomatically or literally in the given context, returning 'i' if the expression is being used idiomatically or 'l' if literally. Expression: {<PIE>}. Context: {<target sentence>}.  Only return one letter (i or l).''

For the FLUTE entailment task, we use: 

``Disambiguate whether the second sentence follows from the first, returning 'entailment' if it does, and 'contradiction' if not. Sentence 1: {<premise sentence>} Sentence 2: {<hypothesis sentence>}.''

\subsection{Prompt Engineering}
We investigate the effect of several prompt variations on performance for GPT-3.5-turbo on the English SemEval test set. As part of the OpenAI API, there are two prompts: ``system''  and ``user''. We first tried using the system prompt to define the task for the model, but obtained better performance using only the user prompt -- this aligns with the experiences of others that GPT-3.5 often doesn't follow the system prompt well, unlike GPT-4\footnote{\url{https://community.openai.com/t/what-is-the-difference-between-putting-the-ai-personality-in-system-content-and-in-user-content/194938}}.

We present our results for this in \Cref{table:results_prompt_engineering}.
Note that variation between runs using the same prompting strategy is high (up to 0.04 F1), which leads to difficulty in discerning the effect of changing the prompt.

\begin{table}[] 
\centering  
\resizebox{\columnwidth}{!}{
\small
\begin{tabular}{ll}
\toprule
 & EN \\
\midrule
Default & 0.739 \\ 
``Expert in language use'' & 0.635 \\
``Expert in language use'' + Idiomatic vs. Compositional & 0.717 \\
``Expert in Idiomatic Language'' & 0.538 \\ 
No ``Only return one letter (i or l).'' & 0.633 \\ 
\bottomrule
\end{tabular}
}
\caption{\label{table:results_prompt_engineering} \centering Results (macro F1) on the English test set of SemEval with GPT-3.5-turbo using prompt engineering.}
\end{table}

Expert impersonation is motivated by work which has shown that prompting LLMs to impersonate domain experts can lead to higher performance \cite{salewski2024context}. As such, we tried two approaches; starting the prompt with ``You are an expert in language use.'' or ``You are an expert in idiomatic language.''. However, we find that neither of these approaches lead to improved performance. Interestingly, replacing the word ``Literal" with ``Compositional" did seem to have a positive effect.
We found that removing the instruction to explicitly return only one letter (`i' or `l') led the model to occasionally return other outputs, which causes a drop in performance (as we treat such responses as invalid). For the English subset, this is the case for 3\% of outputs (28 out of 916 examples).

\subsubsection{Language Prompts}

Since SemEval has test data in English, Portuguese, and Galician, we experiment with a) explicitly stating the language of the sentence in the prompt, and b) translating the prompt using a commercial machine translation tool. We perform this analysis for GPT-3.5-turbo, Gemini 1.0 Pro, and Flan-T5-XXL, with results shown in \Cref{table:results_multilingual_prompt}. 

\begin{table}[h!] 
\centering  
\resizebox{\columnwidth}{!}{
\small
\begin{tabular}{lcccccc}
\toprule
 & \multicolumn{2}{c}{GPT-3.5-turbo} & \multicolumn{2}{c}{Gemini 1.0}  & \multicolumn{2}{c}{Flan-T5-XXL} \\
 & PT & GL & PT & GL & PT & GL\\
\midrule
Default & 0.553 & 0.587 & 0.582 & 0.604 & 0.464 &0.411 \\
Language Prompt & 0.554 & 0.604 &  0.561 & 0.640 & 0.479 &0.457\\
Translated & 0.541 & 0.512 & 0.549 & 0.665 & 0.573 &0.477\\
\bottomrule
\end{tabular}
}
\caption{\label{table:results_multilingual_prompt} \centering GPT 3.5-turbo, Gemini 1.0, and Flan-T5-XXL results for Portuguese and Galician on SemEval using multilingual prompts.}
\end{table}

For Gemini 1.0 Pro and Flan-T5-XXL we see performance improvement for Galician under both of these approaches, with higher performance when translating the prompt.  We hypothesise that both English and Portuguese are likely well-represented in the model training data, and LLMs in general work well in multilingual settings \cite{shi2022language}. However, Galician is likely to be both rare and potentially confused with Portuguese when the language is not specified, or when there is less text in that language available in the prompt. It would be interesting to experiment further with similar language pairs.

Not shown here is that we recorded reduced performance for English across all three models when specifying the language in the prompt (0.739 to 0.674 for GPT-3.5-turbo, 0.771 to 0.732 for Gemini 1.0 Pro, 0.716 to 0.706 for Flan-T5-XXL). It is possible that additional prompt tokens specifying the language may act as a `distractor' when it is the \textit{de facto} default, and the nature of the generative models means that we can anticipate variation in responses to identical prompts.

\subsection{Few-shot Prompting}
The ``one-shot'' setting of SemEval 2022 Task 2a (in which further examples of the target PIE in context are made available) allows for the investigation of passing examples to the model through the prompt. We thus experiment with doing so for GPT-3.5-turbo, Gemini 1.0 and Flan-T5-XXL. We try two configurations: passing one example per PIE (one-shot), and passing all the examples that are available in the dataset (few-shot)\footnote{Where available, the one-shot training data has one idiomatic example for each PIE, and one literal example. However, for some PIEs just one of these is present.}. These results are shown in \Cref{table:results_one_shot}.

\begin{table}[h!] 
\centering  
\resizebox{\columnwidth}{!}{
\small
\begin{tabular}{llllll}
\toprule
 Model & Setting & EN & PT & GL & All \\
\midrule
\multirow{3}{*}{Gemini Pro 1.0} & Zero-shot & \textbf{0.766} & 0.590 & 0.600 & 0.672 \\ 
& One-shot & 0.706 & 0.625 & 0.711 & 0.688  \\
& Few-shot& 0.685 & \textbf{0.642} & \textbf{0.745} & \textbf{0.693} \\
\midrule
\multirow{3}{*}{GPT-3.5-turbo} & Zero-shot & \textbf{0.739} & \textbf{0.563} & \textbf{0.579} & \textbf{0.645} \\
& One-shot & 0.645 & 0.542 & 0.553 & 0.594  \\ 
& Few-shot & 0.686 & 0.545 & 0.566 & 0.614 \\
\midrule
\multirow{3}{*}{Flan-T5-XXL}& Zeroshot &0.629 &0.464 &0.411 &0.514 \\
& Oneshot &0.810 &0.665 &0.732 &0.749 \\
& Fewshot &\textbf{0.845} &\textbf{0.713} &\textbf{0.828} &\textbf{0.805} \\
\midrule
\textit{Best} & Zero-shot & 0.964 & 0.894 & 0.937 & 0.939 \\
\bottomrule
\end{tabular}
}
\caption{\label{table:results_one_shot} \centering Results on SemEval using few-shot prompting.}
\end{table}

Interestingly, the impact of few-shot prompting varies across the models. Flan-T5-XXL benefits the most from this, with stark and consistent performance improvements across the three settings and across all three languages -- the overall F1 jumps from 0.580 in the Zero Shot setting to 0.805 in the Few Shot setting.

Further to this we analyse the performance of all size Flan-T5 models, and present a heatmap illustrating the impacts on performance stemming from zero-shot and few-shot scenarios in \Cref{tab: "prompting results effects heatmap" }. 

\begin{table}[h!]\centering
\scriptsize
\begin{tabular}{lllllll}\toprule
&Small &Base &Large &XL &XXL \\\midrule
Oneshot (EN) &\cellcolor[HTML]{9ed8bc}0.432 &\cellcolor[HTML]{eea9a3}0.079 &\cellcolor[HTML]{fbeceb}0.199 &\cellcolor[HTML]{c7e9d8}0.348 &\cellcolor[HTML]{f9e3e1}0.182 \\
Oneshot (PT) &\cellcolor[HTML]{b3e1ca}0.388 &\cellcolor[HTML]{e7837a}0.011 &\cellcolor[HTML]{fefcfc}0.227 &\cellcolor[HTML]{fefdfc}0.228 &\cellcolor[HTML]{fbeeec}0.202 \\
Oneshot (GL) &\cellcolor[HTML]{70c59c}0.526 &\cellcolor[HTML]{eb9891}0.049 &\cellcolor[HTML]{eb9a94}0.053 &\cellcolor[HTML]{f9e4e2}0.185 &\cellcolor[HTML]{d4eee1}0.321 \\
Oneshot (ALL) &\cellcolor[HTML]{99d6b8}0.443 &\cellcolor[HTML]{ec9b94}0.054 &\cellcolor[HTML]{f7d8d5}0.162 &\cellcolor[HTML]{f0f9f4}0.264 &\cellcolor[HTML]{fefffe}0.235 \\
\midrule
Fewshot (EN) &\cellcolor[HTML]{75c79f}0.516 &\cellcolor[HTML]{e67c73}-0.003 &\cellcolor[HTML]{cfecdd}0.332 &\cellcolor[HTML]{acdec5}0.404 &\cellcolor[HTML]{fdf6f6}0.217 \\
Fewshot (PT) &\cellcolor[HTML]{b2e0c9}0.391 &\cellcolor[HTML]{e67d74}0.000 &\cellcolor[HTML]{f0b1ac}0.093 &\cellcolor[HTML]{e6f5ee}0.285 &\cellcolor[HTML]{f7fcfa}0.249 \\
Fewshot (GL) &\cellcolor[HTML]{57bb8a}0.576 &\cellcolor[HTML]{e67d74}0.000 &\cellcolor[HTML]{f4cac6}0.137 &\cellcolor[HTML]{c4e7d6}0.354 &\cellcolor[HTML]{a5dbc1}0.417 \\
Fewshot (ALL) &\cellcolor[HTML]{82cda8}0.489 &\cellcolor[HTML]{e67c73}-0.001 &\cellcolor[HTML]{fefcfc}0.227 &\cellcolor[HTML]{c5e8d7}0.352 &\cellcolor[HTML]{e3f4eb}0.291 \\
\bottomrule
\end{tabular}
\caption{\label{tab: "prompting results effects heatmap" } \centering Enhancements in Macro F1 scores (positive values) and declines (negative values) when compared to the performance in zero-shot conditions across all Flan-T5 models.}
\end{table}

The smallest models benefited the most from seeing one or more examples before inference. In the best cases, performance in English improved by 0.432 in the one-shot setting and 0.516 in the few-shot setting. Interestingly, few-shot prompting can be seen to improve performance across Portuguese and Galician examples in all model settings, apart from T5-FLAN-Base and Large where there is little, or no improvement. 
It appears that Flan-T5-Base seems to be least improved by prompting with examples, with a negative effect on performance in few-shot prompting settings. In the one-shot setting, improvement in model performance is minor. 
The Large, XL and XXL models also benefited from one- and few-shot prompting, with Flan-T5-XL seeing the most performance enhancement. It appears that whilst models follow ``bigger is better" in zero-shot settings, they do not necessarily follow this pattern under one/few-shot prompting. In fact, the best performance in the few-shot setting is with T5-Small, which at only 80M parameters achieves an overall F1 of 0.821, the best performance of any of the models we have evaluated in this paper. This is in significant contrast to performance on MAGPIE and FLUTE, where zero-shot performance is very low. The model is likely learning some artefacts from the data such as predicting only one label for a given PIE in the SemEval dataset.

Gemini 1.0 Pro also achieves consistent (though smaller) performance improvements from Zero Shot to One Shot to Few Shot, but the performance for English reverses this pattern. We also see a big jump in performance between Zero Shot and One Shot for Galician, which we again attribute to the rarity of this language and its similarity with Portuguese.

GPT-3.5-turbo is hindered by providing examples. The reasons for this are unclear, but this may be linked to the inability shown by GPT-3.5 to follow system prompts. If the model is not successfully following longer prompts then they may effectively introduce noise and lead to worse performance, as we saw when comparing results with and without system prompts.

\section{Discussion}

\subsection{Task Labelling}

The majority of the models we examined achieved high performance on the FLUTE dataset. We attribute this to the nature of FLUTE's evaluation being distinct from MAGPIE and SemEval. For the latter two, the model is asked to label `idiomatic' or `literal' use of a given idiom, whereas, in the FLUTE STS task, the model is required to pick out the contradiction or entailment relationship between two sentences.

This means that a model might not necessarily require `knowledge' of the target idiom to succeed, but could determine the relationship between the two sentences from other information, as facilitated by contextualised embeddings \cite{boisson_constructionartifacts}. Moreover, the model is likely to have encountered similar tasks during its pre-training. Flan-T5 models are instruction-refined versions of T5 \citep{raffel2020exploring,chung2022scaling}, that have undergone exposure to over 1000 tasks during its fine-tuning process alone. Among these tasks are evaluations of entailment and contradiction judgments, akin to FLUTE, such as SNLI \citeplanguageresource{snli}, MNLI \citeplanguageresource{mnli}, CB \citeplanguageresource{CB} and numerous other reasoning tasks \cite[for details see][]{raffel2020exploring,chung2022scaling}.



\subsection{Practicalities}

In contrast with fine-tuned classification models, as prompted models are capable of open-ended generation, they may not output a response in the format requested. While the output may be readily interpretable by a human reader, this is not practical when evaluating large numbers of responses. Prompting for specific formats is easier for models which have undergone more instruction tuning \cite{ouyang2022training, rafailov2023direct}, and is a key reason why the Mistral-7B model outperforms the Llama2 7B variant.

Prompted, generative models produce outputs which are subject to variation when they are repeatedly given the same prompt. While the user may have some control over this behaviour through `temperature' parameters, this variability is inherent to generative models. When converting the outputs of such models to a labelling decision, this variability will also affect the results.

Despite their generally higher performance than the local models and their advantages when it comes to prototyping, there are a number of considerations specific to SaaS models which may be significant. These include:

\begin{enumerate}
    \item Cost -- The larger models have a higher per-1000-tokens cost, which may lead to some evaluations being cost-prohibitive. Evaluating GPT-4 on the (relatively small) SemEval test set, for example, costs \$11. Running evaluation on this model, especially across multiple runs for prompt tuning, etc. may potentially price out researchers with lower budgets.
    
    \item Safety Features -- Commercial SaaS models frequently include features designed to limit models and users' capability to process or generate content which may cause harm. These features may also impact on researchers' ability to use the tools, as they produce what are effectively false positives.
    For example, when using the VertexAI API for experiments with Gemini Pro, the API consistently refused to generate responses for a small number of prompts. These included certain contexts for the expression \textit{street girl} which referred to prostitution or sexualization, but also the FLUTE sentence pair ``Your brother is mature and behaves in an adult manner. Your brother is a big baby." for the expression \textit{to be a big baby}\footnote{Replacing the word `adult' with `grown-up' convinced the service to generate a response.}. We treat any such responses as incorrect in our statistics.
    
    \item Service Changes -- Changes to the underlying model can be made by the third party at any time, and can significantly impact the performance of the models and the consistency of results. Whilst undertaking this work the default gpt-3.5-turbo model changed from one released in June 2023, to one released in January 2024.
    
    \item Rate limits -- For larger datasets, the rate limits of commercial APIs can become an issue. As it is still not fully released, for a significant amount of time during the creation of this work, the daily rate limit for GPT-4-turbo was lower than the number of tokens in MAGPIE, which prevented us from completing any evaluation runs for this model and dataset combination.
\end{enumerate}






\section{Conclusion}
In this work we have evaluated the performance of various large language models on three idiomaticity datasets (SemEval 2022 Task 2a, FLUTE, and MAGPIE). We have investigated locally-run models up to 13B parameters, as well as significantly larger models (GPT-3.5, GPT-4, and Gemini 1.0 Pro) accessed through commercial APIs. We perform an extensive analysis of the impact of several factors on performance; model size, prompt engineering and few-shot prompting. In addition, we discuss considerations for practitioners wishing to use these models in their own work, with emphasis on cost and practicalities such as the variability of outputs and the impacts of decisions made by the companies operating these services. 
Our overall findings are as follows: 1) LLMs at the highest scale are able to achieve competitive results for idiomaticity detection, and performance on FLUTE in particular seems to have saturated, but these general models do not match the performance of (much-smaller) encoder models fine-tuned for the specific idiomaticity detection tasks of SemEval and MAGPIE. 2) The performance of prompted, generative LLMs seems to scale consistently with parameter count for these datasets, indicating the potential of even bigger models to achieve further increases in performance. 3) While they are based on a relatively small set of examples, our experiments with multilingual models suggest that performance gains can be obtained by specifying the target language, translating prompts and by providing examples. However, the efficacy of these modifications depends on the model used and the language in question; they appear to harm performance for English (which is, presumably, the most-represented language in the model training regimens) while producing the largest benefit for the much rarer Galician.

\section*{Acknowledgments}
We would like to thank the following sponsors for supporting this work:
\begin{itemize}
    \item The Healthy Lifespan Institute (HELSI) at The University of Sheffield, UK EPSRC grant EP/T517835/1,  
    \item The Centre for Doctoral Training in Speech and Language Technologies (SLT) and their Applications funded by UK Research and Innovation [grant number EP/S023062/1], and
    \item UK EPSRC grant EP/T02450X/1, by the Turing Institute Fellowship and by the UniDive COST Action.
\end{itemize}

\nocite{*}
\section{Bibliographical References}\label{sec:reference}

\bibliographystyle{lrec-coling2024-natbib}
\bibliography{lrec-coling2024-example}

\section{Language Resource References}
\label{lr:ref}
\bibliographystylelanguageresource{lrec-coling2024-natbib}
\bibliographylanguageresource{languageresource}

\end{document}